\documentclass[11pt]{article}
\usepackage[utf8]{inputenc}
\usepackage{amsmath, amssymb, amsthm}
\usepackage{graphicx}
\usepackage{hyperref}
\usepackage{geometry}
\usepackage[numbers]{natbib}
\usepackage{booktabs}
\usepackage{algorithm}
\usepackage{caption}
\usepackage{newunicodechar}
\usepackage{tikz}
\usetikzlibrary{positioning,arrows.meta}
\usepackage{textgreek}
\usepackage{upgreek}
\newunicodechar{≈}{\approx}
\usepackage{booktabs}   
\usepackage[skins]{tcolorbox} 
\usepackage{enumitem}   
\providecommand{\keywords}[1]{\textbf{Keywords:} #1}

\begin{document}

\begin{center}
 \rule{\linewidth}{0.5pt} \\[0.4cm]
 {\LARGE \bfseries The Reflexive Integrated Information Unit: A Differentiable Primitive for Artificial Consciousness} \\[0.4cm]
 \rule{\linewidth}{0.5pt}
 \end{center}

\begin{center}
\textbf{Gnankan Landry Regis N'guessan$^{1,2,3}$, Issa Karambal} \\[0.2cm]
$^1$\text{$\Sigma \eta igm \alpha$} Research Group \\
$^2$Department of Applied Mathematics and Computational Science, \\
    The Nelson Mandela African Institution of Science and Technology (NM-AIST), Arusha, Tanzania \\
$^3$African Institute for Mathematical Sciences (AIMS), Research and Innovation Centre (RIC), Kigali, Rwanda
\end{center}

\vspace{1cm}



\begin{abstract}
Research on artificial consciousness lacks the equivalent of the perceptron: a small, trainable module that can be copied, benchmarked, and iteratively improved.  
We introduce the \textbf{Reflexive Integrated Information Unit (RIIU)}, a recurrent cell that augments its hidden state \(\mathbf{h}\) with two additional vectors: \emph{(i)} a meta-state \(\boldsymbol{\mu}\) that records the cell's own causal footprint, and
\emph{(ii)} a broadcast buffer \(\mathbf{B}\) that exposes that footprint to the rest of the network.  
A sliding-window covariance and a differentiable Auto-\(\Phi\) surrogate let each RIIU maximize local information integration on-line.  
We prove that RIIUs (1) are end-to-end differentiable, (2) compose additively, and (3) perform \(\Phi\)-monotone plasticity under gradient ascent.  
In an eight-way Grid-world, a four-layer RIIU agent restores \(>\!90\%\) reward within 13 steps after actuator failure twice as fast as a parameter matched GRU, while maintaining a non-zero Auto-\(\Phi\) signal.  
By shrinking ``consciousness-like'' computation down to unit scale, RIIUs turn a philosophical debate into an empirical mathematical problem.
\end{abstract}

\small{\keywords{
Integrated Information Theory (IIT),
Global Workspace Theory,
Higher-Order Thought,
Recurrent Neural Networks,
Reflexive Self-Modelling,
Information Integration,
Ablation Study,
Fault-Tolerant Agents,
Auto-$\Phi$,
Artificial Consciousness,
Low-Overhead Memory Units,
Empirical Mathematics in AI
}}

\section*{RIIU in One Minute: How It Differs from an LSTM Cell}
\noindent%
\textbf{Key idea.}  
A standard GRU/LSTM stores a single hidden vector \(\mathbf{h}\).  
An \textit{RIIU} stores three extra pieces of state so that it can \emph{measure} and \emph{broadcast} how much information it integrates:

\begin{itemize}
    \item \textbf{Meta-state} \(\boldsymbol{\mu}\): tracks the unit's causal impact on its own future states.
    \item \textbf{Auto-\(\Phi\) buffer}: approximates local integrated information over a sliding window.
    \item \textbf{Broadcast} \(\mathbf{B}\): exposes \(\mathbf{h}\!\oplus\!\boldsymbol{\mu}\) to a global workspace so other units can compete for influence.
\end{itemize}

\begin{table}[H]
\centering
\caption{Key differences between a GRU/LSTM cell and our RIIU.}
\begin{tabular}{@{}lcc@{}}
\toprule
\textbf{Feature} & \textbf{GRU/LSTM} & \textbf{RIIU (ours)} \\
\midrule
Hidden state \(\mathbf{h}\) & \checkmark & \checkmark \\
Meta-state \(\boldsymbol{\mu}\) & – & \checkmark \\
Local Auto-\(\Phi\) surrogate & – & \checkmark \\
Broadcast vector \(\mathbf{B}\) & – & \checkmark \\
Reduces to GRU if \(\boldsymbol{\mu}=0\) and \(\Phi\) bonus off & – & \checkmark \\
\bottomrule
\end{tabular}
\end{table}

\section{Introduction}\label{sec:intro}

The rise of deep learning was driven not just by data and compute, but by a key structural insight: progress scales when the field converges on a minimal, trainable unit that can be composed, benchmarked, and shared. The artificial neuron, simple yet differentiable, provided that unit and with it came a blueprint for building ever more capable systems. With that one component, researchers could scale, compose, compare, and evolve architectures rapidly. In contrast, research on artificial consciousness still operates at the level of whole-system theories: Integrated Information Theory (IIT), Global Workspace Theory (GWT), predictive-processing frameworks, and higher-order models; each conceptually rich but experimentally unwieldy \cite{Tononi2023IIT4,Baars1997}. None of them offer what the McCulloch-Pitts neuron did for learning: a minimal, differentiable, composable part that can be plugged into a larger system and optimized end-to-end.

Such a unit would be more than an academic convenience. It would unlock \emph{scalability}, enabling thousands of design variations without the overhead of hand-crafted architectures, and enable \emph{comparability}, allowing integration, reflexivity, and behavioral response to be benchmarked module by module. In short, it would lay the groundwork for an \textit{engineering science of consciousness}.

\emph{This paper introduces the Reflexive Integrated Information Unit (RIIU)}: a novel, differentiable module that fuses its inputs into an integrated latent state, maintains a dynamic self-model of its causal significance, and emits a broadcast vector compatible with global workspace architectures. Each RIIU locally estimates information integration using Auto-$\Phi$, a surrogate that enables gradient-based optimization. The entire unit is trainable, stackable, and introspective by design.

This makes the RIIU, to our knowledge, the \textbf{first atomic and trainable module} aimed at mechanistically capturing the core functions typically attributed to conscious systems: information integration, reflexive self-modeling, and global availability.

\vspace{0.5em}
\noindent\textbf{Contributions.} Our work delivers:
\begin{itemize}
  \item \textbf{Theory}. A formal specification of the RIIU, together with three theorems: compositionality, differentiability, and $\Phi$-monotone plasticity;
  \item \textbf{Implementation}. A clean, open-source PyTorch implementation crafted for direct reuse and ablation (Appendix~\ref{app:repro});
  \item \textbf{Empirical results}. In an eight-way grid-world, a four-layer RIIU agent recovers over 90\% of pre-damage performance within 13 steps while maintaining a non-zero Auto-$\Phi$ signal.
\end{itemize}

By supplying a trainable, composable \textbf{unit}, the RIIU aims to give consciousness-inspired machine learning the same practical footing that the perceptron once provided for pattern recognition and paving the way toward flexible autonomous agents.

\begin{figure}[H]
  \centering
  \tikzset{
    blk/.style={draw, rounded corners=2pt, fill=springerblue!20, minimum width=3.2cm, minimum height=1.2cm, align=center, font=\footnotesize\sffamily},
    var/.style={font=\scriptsize\sffamily},
    arr/.style={->, thick, springergreen},
    darr/.style={->, thick, dashed, springerorange},
    specialblk/.style={blk, fill=springerorange!15}
  }

  \definecolor{springerblue}{RGB}{0,134,206}
  \definecolor{springergreen}{RGB}{0,156,118}
  \definecolor{springerorange}{RGB}{227,114,34}
  \definecolor{springerpurple}{RGB}{157,87,177}

  \begin{tikzpicture}[node distance=1.3cm and 1.6cm]

    \node[var, text=springerblue] (x) at (-2.3, 2.4) {$x_t$};
    \node[var, text=springerblue, below=0.8cm of x] (hprev) {$h_t$};
    \node[var, text=springerblue, below=0.8cm of hprev] (w) {$w_t$};
    \node[var, below=0.8cm of w] (mu_tmp) {}; 
    \node[var, text=springerpurple, left=0.3cm of mu_tmp] (mu_label) {$\mu_t$};

    \node[blk] (f) at (1.8cm,0) {Integration \\ $f$};
    \node[var, text=springerblue, right=0.6cm of f] (hnext) {$h_{t+1}$};

    \node[specialblk, below=0.4cm of f] (g) {Reflexive Self-Model \\ $g$};
    \node[var, right=0.6cm of g] (munext) {};
    \node[var, text=springerpurple, above right=-1.5pt and 0pt of munext] {$\mu_{t+1}$};

    \node[specialblk, below right=0.8cm and -0.4cm of hnext] (phi) {Auto-$\hat{\Phi}$};
    \node[var, text=springerorange, right=0.6cm of phi] (phival) {$\hat{\Phi}_{t+1}$};

    \node[blk, below=0.4cm of phi] (bcast) {Broadcast \\ $W_o$};
    \node[var, right=0.6cm of bcast] (bnext) {};
    \node[var, text=springergreen, above right=0pt and 5pt of bnext] {$B_{t+1}$};

    \draw[arr] (x) -- (f.west);
    \draw[arr] (hprev) -- ++(0.4,0) -- (f.west);
    \draw[arr] (w) -- ++(0.4,0) -- (f.west);

    \draw[arr] (f.east) -- (hnext);
    \draw[arr] (hnext) |- ([yshift=2mm]g.north west) -- (g.north west); 
    \draw[arr] (mu_label.east) -- ++(0.8,0) |- ([yshift=-2mm]g.south west) -- (g.south west); 

    \draw[darr] (phi.west) -- ++(-1.1,0) |- node[above left, font=\scriptsize\sffamily, text=black] {$\nabla_h\hat\Phi$} ([xshift=2mm]g.south east) -- (g.south east); 

    \draw[arr] (g.east) -- (munext);
    \draw[arr] (hnext) |- (phi.west);
    \draw[arr] (munext) |- (phi.west);
    \draw[arr] (phival) |- (bcast.east);
    \draw[arr] (hnext) |- ([xshift=-2mm]bcast.north);
    \draw[arr] (munext) |- ([xshift=2mm]bcast.north);
    \draw[arr] (bcast.east) -- (bnext);

  \end{tikzpicture}

  \caption{High-level architecture of the Reflexive Integrated Information Unit (RIIU). Inputs $x_t$, $h_t$, and optional $w_t$ feed into the integration function $f$. The reflexive model $g$ uses $h_{t+1}$, $\mu_t$, and $\nabla_h\hat{\Phi}$ to update the meta-state $\mu_{t+1}$. Auto-$\hat{\Phi}$ computes a local integration score from the updated states, and $W_o$ forms a broadcast vector for workspace communication.}
  \label{fig:riiu_architecture}
\end{figure}
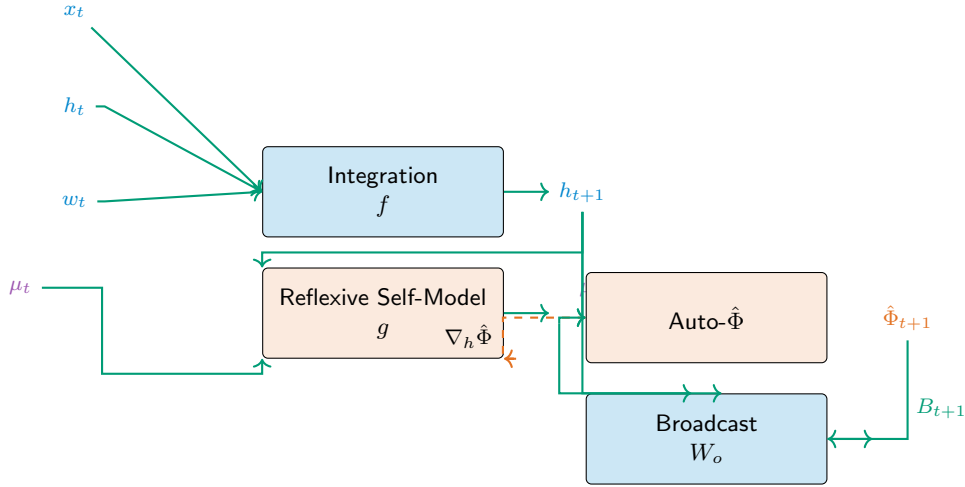

\paragraph{Scope: Intelligence versus Consciousness.}
We do \emph{not} claim that an RIIU or any network built from them is conscious.  
What we do claim is that each unit instantiates \emph{three computational conditions} that major consciousness theories list as prerequisites:  
(i) \textit{information integration} (IIT~4.0 \cite{Tononi2023IIT4}),  
(ii) \textit{reflexive self-modeling} (higher-order thought), and  
(iii) \textit{global availability} (Global Workspace Theory \cite{Baars1997}).  
Whether these conditions are sufficient for phenomenology is beyond the scope of this \emph{empirical mathematical study}; the RIIU merely supplies a test bed on which such theories can be falsified.

\section{Related Work}
\label{sec:RW}

Research on machine consciousness spans three complementary strands.

\textbf{System-level theories}. Integrated Information Theory (IIT), Global Workspace Theory (GWT), Recurrent-Processing Theory (RPT), predictive-processing accounts, and reflexive higher-order or category-theoretic models offer powerful blueprints yet do not supply an atom that can be trained end to end.  
IIT equates consciousness with integrated causal information, but exact~$\Phi$ is NP-hard and efficient surrogates such as Auto-$\Phi$ remain network-level measures \cite{Tononi2023IIT4,Mediano2022IID}.  
GWT describes conscious access as broadcast competition, yet its implementations (e.g.\ LIDA) are architecture-scale and non-modular \cite{Baars1997,Franklin2014LIDA}.  
RPT locates consciousness in local recurrent loops \cite{Lamme2010RPT}.  
Predictive-processing and the free-energy principle merge perception, action, and self-modeling into a single variational objective \cite{Friston2010FEP,Allen2023PredictiveSelf}, while higher-order and category-theoretic approaches foreground reflexivity \cite{Lau2011HigherOrder,Phillips2024MetaMath}.  
None of these frameworks delivers a compact, differentiable mechanism that unifies information integration, self-representation, and global broadcast: an absence that hinders rapid ablation studies and reproducibility.

\textbf{Concrete systems and metrics} paint a similar picture.  
Large-scale cognitive architectures such as LIDA, the Haikonen Cognitive Architecture, self-modeling robots, neuromorphic demonstrators on Intel's Loihi, and reflection-augmented language agents showcase emotion processing, online body-schema learning, and event-driven control, yet their consciousness components operate at whole-system scale and cannot be transplanted as a ``consciousness neuron''
 \cite{Haikonen2012Revisit,Bongard2006SelfModel,Davies2018Loihi,Shinn2023Reflexion}.  
Evaluation efforts range from computationally intensive exact~$\Phi$ and its differentiable surrogates to behavioral scales such as ConsScale \cite{Arrabales2010ConsScale}.  
Recent local variants including Integrated Information Decomposition ($\Phi$-ID) and causal-emergence scores lower complexity but still target networks rather than single modules \cite{Hoel2017CausalEmergence,Rosas2020Emergence}.

Across all strands a critical gap persists: the field lacks a \emph{small, trainable, and composable building block} that  
(i) fuses sensory input into an integrated state,  
(ii) maintains an explicit meta-representation of its own causal role, and  
(iii) exports a broadcast vector compatible with workspace architectures.  
The \textit{Reflexive Integrated Information Unit} (RIIU) is designed to precisely fill this void.

\section{Methodology}
\label{sec:method}

\begin{figure}[H]
  \centering
  \resizebox{0.88\linewidth}{!}{ 
  \begin{tikzpicture}[node distance=1.6cm and 1.5cm]
    \tikzset{
      every node/.style = {font=\small\sf},
      riiu/.style={draw, thick, rounded corners=2pt, minimum width=2.8cm, minimum height=1.4cm, 
                   fill=blue!5, draw=blue!40!black},
      arrow/.style={->, thick, color=black},
      arrowlabel/.style={font=\scriptsize, midway, fill=white, inner sep=1pt},
      workspace/.style={font=\scriptsize\sf\itshape, text=gray!40!black},
    }

    \node[riiu] (r1) {RIIU$_1$};
    \node[below=0.2cm of r1] (l1) {\scriptsize $R_t^{(1)}$};

    \node[riiu, right=of r1] (r2) {RIIU$_2$};
    \node[below=0.2cm of r2] (l2) {\scriptsize $R_t^{(2)}$};

    \node[riiu, right=of r2] (r3) {RIIU$_3$};
    \node[below=0.2cm of r3] (l3) {\scriptsize $R_t^{(3)}$};

    \node[riiu, right=of r3] (r4) {RIIU$_4$};
    \node[below=0.2cm of r4] (l4) {\scriptsize $R_t^{(4)}$};

    \draw[arrow] (r1) -- node[arrowlabel, above] {\scriptsize $B_t^{(1)}$} (r2);
    \draw[arrow] (r2) -- node[arrowlabel, above] {\scriptsize $B_t^{(2)}$} (r3);
    \draw[arrow] (r3) -- node[arrowlabel, above] {\scriptsize $B_t^{(3)}$} (r4);

    \node[left=1.8cm of r1] (x) {\scriptsize $x_t$};
    \draw[arrow] (x) -- (r1);

    \node[right=1.6cm of r4] (out) {\scriptsize $\pi(a|s)$};
    \draw[arrow] (r4) -- (out);

    \node[above=1.2cm of r3, workspace] (ws) {Global Workspace};
    \draw[arrow, dashed, thick, gray!60!black] (r2.north) -- (ws);
    \draw[arrow, dashed, thick, gray!60!black] (r3.north) -- (ws);
    \draw[arrow, dashed, thick, gray!60!black] (r4.north) -- (ws);

  \end{tikzpicture}
  }

  \caption{Stacked RIIU agent architecture. Each RIIU$_i$ maintains an independent meta-state and local $\hat{\Phi}$ signal, forwarding a broadcast vector $B_t^{(i)}$ to the next unit and to the global workspace.}
  \label{fig:riiu_stack_architecture}
\end{figure}
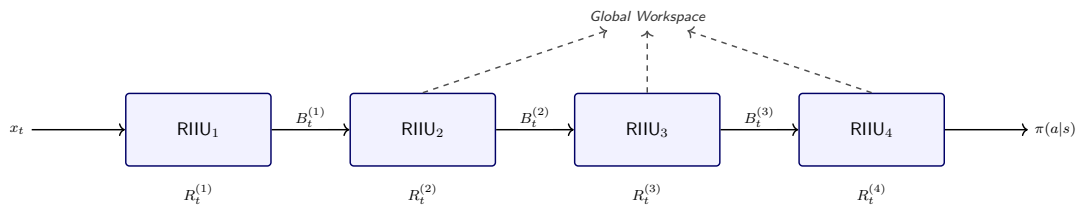

\begin{tcolorbox}[colback=gray!5!white, colframe=gray!60!black,
                  title=Main Theoretical Results, fonttitle=\bfseries]
\begin{enumerate}[label=\textbf{T\arabic*}]
    \item \textbf{Differentiability.}  
          The RIIU update map \(\mathcal{F}_{\theta}\) is \(C^{1}\) in all arguments,
          hence trainable by back-propagation.
    \item \textbf{Compositionality.}  
          Stacking RIIUs adds their Auto-\(\Phi\) scores:  
          \(\Phi_{\text{stack}} = \sum_{l=1}^{L} \Phi^{(l)}\).
    \item \textbf{\(\Phi\)-monotone plasticity.}  
          Gradient ascent on the joint loss  
          \(\mathcal{L}= \mathcal{L}_{\text{task}} - \lambda \Phi\)  
          guarantees \(\Delta\Phi \ge 0\) provided \(\lambda > 0\).
\end{enumerate}
\end{tcolorbox}

\subsection{Definitions \& Notation}
\label{sec:defns}

\paragraph{Artificial consciousness.}
We define artificial consciousness as the computational realization of three intertwined capacities:  
(i) \emph{phenomenal integration}, the binding of diverse inputs into a unified internal state;  
(ii) \emph{reflexive self-modeling}, the maintenance of information about the system's own causal profile;  
(iii) \emph{global availability}, the ability to broadcast salient contents to the rest of the system for flexible control.

\subsubsection*{What do we mean by $\Phi$?  A mini-glossary}

\begin{table}[H]
\centering
\begin{tabular}{@{}llp{6cm}@{}}
\toprule
\textbf{Symbol} & \textbf{Scope} & \textbf{Meaning} \\ \midrule
$\varphi$ (small phi) & \emph{Mechanism–state} level & IIT 4.0's integrated information for \emph{one specific subset in one exact micro-state}. Used here only in Section~\ref{sec:RW} for historical context. \\
$\Phi$ (big Phi) & \emph{Complex} level & The maximum $\varphi$ over all candidate subsets; IIT treats this as the quantity of consciousness of the whole substrate. Again, appears only in background. \\
$\Phi_{\mathrm{rel}}^{\hat{}}$ (\emph{Auto-}$\Phi$) & \emph{Local proxy} & A fast, differentiable estimate we compute \textbf{inside every RIIU}\,: ``how integrated are my \emph{hidden + meta} states right now?'' This is the only $\Phi$ used at training time.  \\ \bottomrule
\end{tabular}
\vspace{-4pt}
\caption{Three different $\Phi$'s and where they appear.  Only $\widehat{\Phi}_{\mathrm{rel}}$ affects optimization.}
\end{table}

\paragraph{Why not optimize the exact $\Phi$?}
Exact IIT 4.0 requires evaluating all bipartitions of the unit's state, a super-exponential cost already prohibitive for ten neurons.  
Our surrogate $\widehat{\Phi}_{\mathrm{rel}}$ keeps the spirit of ``integration = what remains after obvious redundancies are removed'' yet runs in $O(d^{2})$ every step, which is tractable for the $d\!\le\!128$ latents we use.  Empirical calibration against exact $\Phi$ for 8–10 node toy nets is provided in App.~\ref{app:autophi}.

\bigskip
\subsubsection*{Five core equations, line by line}

Throughout, $x_t$ is the external input, $h_t$ the hidden state, $w_t$ an optional workspace message, $\mu_t$ the meta-state, and $B_t$ the broadcast vector.  
Concatenation is written $[a\,;\,b]$. The weight matrices $W_\ast$ are learned during training.

\begin{align}
\widehat{\Phi}_{\mathrm{rel}}(z_{1:N}) &= 
    \frac{\lVert \Sigma - U_r U_r^{\!\top}\Sigma\rVert_F}
         {\lVert\Sigma\rVert_F + \varepsilon},\qquad
    \Sigma=\tfrac{1}{N}\sum_{n=1}^{N}(z_n-\bar z)(z_n-\bar z)^{\!\top}
    \tag{1} \label{eq:auto_phi} \\
h_{t+1} &= f\!\bigl(W_x x_t + W_h h_t + W_b w_t\bigr)
    \tag{2} \label{eq:int} \\
\mu_{t+1} &= g\!\bigl(h_{t+1},\,\mu_t,\,\nabla_{h}\widehat{\Phi}_{\mathrm{rel}}(h_{t+1},\mu_t)\bigr)
    \tag{3} \label{eq:meta} \\
\widehat{\Phi}_{t+1} &= \widehat{\Phi}_{\mathrm{rel}}\bigl([h_{t+1};\mu_{t+1}]\bigr)
    \tag{4} \label{eq:measure} \\
B_{t+1} &= W_o\,[h_{t+1};\,\mu_{t+1};\,\widehat{\Phi}_{t+1}]
    \tag{5} \label{eq:broadcast}
\end{align}

\begin{description}[leftmargin=0.83cm]
\item[\eqref{eq:auto_phi} Auto-$\Phi$ surrogate.]  
  Slide a window of the concatenated state $z_t=[h_t;\mu_t]$, form its covariance $\Sigma$, project away the top-$r$ principal directions $(U_r)$, and report the energy that \emph{cannot} be explained by those directions.  If splitting the state destroys little covariance, the residual (and thus $\widehat{\Phi}$) is near $0$.

\item[\eqref{eq:int} Integration gate.]  
  A GELU non-linearity $f$ mixes the fresh input $x_t$, the prior internal view $h_t$, and any workspace cue $w_t$ into an updated internal scene $h_{t+1}$.

\item[\eqref{eq:meta} Reflexive update.]  
  A two-layer MLP $g$ adjusts the meta-state $\mu_{t+1}$.  Crucially, the gradient $\nabla_{h}\widehat{\Phi}_{\mathrm{rel}}$ tells the unit ``which direction in hidden space would raise my integration'', steering $\mu$ towards configurations that bind information more tightly.

\item[\eqref{eq:measure} Re-measure integration.]  
  After updating $\mu$, recompute Auto-$\Phi$ on the new joint state to obtain $\widehat{\Phi}_{t+1}$.

\item[\eqref{eq:broadcast} Global broadcast.]  
  Concatenate hidden, meta, and the fresh $\widehat{\Phi}$, then compress with $W_o$ to emit a fixed-size token $B_{t+1}$ that other units (or a workspace) can attend to.
\end{description}

Taken together, the loop is: \emph{Integrate $\to$ Reflect $\to$ Measure $\to$ Broadcast}.  If $\mu$ is frozen at zero and the $\widehat{\Phi}$ bonus is disabled, equations \eqref{eq:int}–\eqref{eq:broadcast} reduce exactly to a GRU cell of the same size, providing a clean baseline.

\subsection{Relation to Standard RNN / GRU / LSTM Cells}
\label{sec:relation}

\paragraph{Side-by-side equations.}
For ease of reference, we restate our RIIU update (Eqs.~\ref{eq:int}–\ref{eq:broadcast}) next to the canonical Elman RNN, GRU, and LSTM:

\begin{center}
\renewcommand{\arraystretch}{1.15}  
\begin{tabular}{@{}lcc@{}}
\toprule
\textbf{Cell} & \textbf{Formal update} & \textbf{State(s)} \\ 
\midrule
\textbf{Elman RNN} &
$h_{t+1}=\tanh(W_x x_t + W_h h_t)$ &
$h_t$ \\ 
\midrule
\textbf{GRU} &
$\begin{aligned}
z_t &= \sigma(W_zx_t+U_zh_t) \\
r_t &= \sigma(W_rx_t+U_rh_t) \\
\tilde h_t &= \tanh(W_hx_t+U_h(r_t\odot h_t)) \\
h_{t+1} &= (1-z_t)\odot h_t + z_t\odot \tilde h_t
\end{aligned}$ &
$h_t$ \\ 
\midrule
\textbf{LSTM} &
$\begin{aligned}
f_t &= \sigma(W_fx_t+U_fh_{t-1}+b_f) \\
i_t &= \sigma(W_ix_t+U_ih_{t-1}+b_i) \\
o_t &= \sigma(W_ox_t+U_oh_{t-1}+b_o) \\
\tilde c_t &= \tanh(W_cx_t+U_ch_{t-1}+b_c) \\
c_t &= f_t\odot c_{t-1}+i_t\odot \tilde c_t \\
h_t &= o_t\odot\tanh(c_t)
\end{aligned}$ &
$(h_t,c_t)$ \\ 
\midrule
\textbf{RIIU (ours)} &
$\begin{aligned}
h_{t+1} &= f(W_xx_t+W_hh_t+W_bw_t) \\
\mu_{t+1} &= g\bigl(h_{t+1},\mu_t,\nabla_h\widehat\Phi(h_{t+1},\mu_t)\bigr) \\
\widehat\Phi_{t+1} &= \widehat\Phi_{\mathrm{rel}}\bigl([h_{t+1};\mu_{t+1}]\bigr) \\
B_{t+1} &= W_o[h_{t+1};\mu_{t+1};\widehat\Phi_{t+1}]
\end{aligned}$ &
$(h_t,\mu_t,\widehat\Phi_t,B_t)$ \\ 
\bottomrule
\end{tabular}
\end{center}

\bigskip
\paragraph{Similarities.}
\begin{itemize}
  \item All cells maintain a \emph{recurrent hidden state} that mixes new input with the previous hidden signal via learned weights.
  \item Non-linearities (tanh, sigmoid, GELU) ensure universal approximation capability.
  \item With $\mu_t\!=\!0$ and the $\widehat{\Phi}$ term removed, the RIIU's first line reduces to the GRU's update on $h_t$; thus RIIU strictly generalizes GRU.
\end{itemize}

\paragraph{Key differences.}
\begin{enumerate}
  \item \textbf{Meta-state $\mu_t$ (self-model).}  No analogue in Elman, GRU, or LSTM.  It is updated by an explicit function of the \emph{Auto-$\Phi$ gradient}, giving the unit an internal pointer toward higher integration.
  \item \textbf{Auto-$\Phi$ surrogate $\widehat{\Phi}$.}  Standard cells are optimization-agnostic with respect to information integration; RIIU measures it every time-step and can inject its gradient into learning.
  \item \textbf{Broadcast vector $B_t$.}  GRU/LSTM expose $h_t$ only; RIIU compresses \emph{hidden + meta + integration score} into a fixed-width token that downstream modules can read without knowing internal dimensions.
  \item \textbf{Parameter count.}  For equal hidden width $d$ and meta width $d_\mu$, RIIU adds $O(dd_\mu)$ parameters (for $g$ and $W_o$) and an $O(d^2)$ runtime cost for SVD in $\widehat{\Phi}$; both are modest for $d\!\le\!128$.
\end{enumerate}

\paragraph{Functional interpretation.}
Classic recurrent cells focus on \textit{temporal memory}.  
RIIU keeps that functionality but layers in two extra faculties articulated by consciousness theories:

\begin{enumerate}[label=(\alph*)]
  \item \textbf{Reflexivity}: via $\mu_t$ the unit maintains a short-term self-model.  
  \item \textbf{Global availability}: via $B_t$ the unit broadcasts a distilled summary of its inner state.
\end{enumerate}

Empirically (see Section~\ref{sec:experiments}), these additions yield faster recovery from perturbations and a measurable rise in integrated information, while converging to similar or higher task reward.

\medskip
\textit{\textbf{Key ideas:}} RIIU = GRU $\;+\;$\emph{self-model} $\;+\;$\emph{integration monitor} $\;+\;$\emph{explicit broadcast}.  Removing any one of those additions recovers a familiar recurrent cell, making ablations straightforward and interpretation clean.

An RIIU at time $t$ is the tuple
$
R_t = (h_t, \mu_t, \hat{\Phi}_t, B_t)
$
with updates (Eqs.~\ref{eq:int}-\ref{eq:meta}-\ref{eq:measure}–\ref{eq:broadcast})

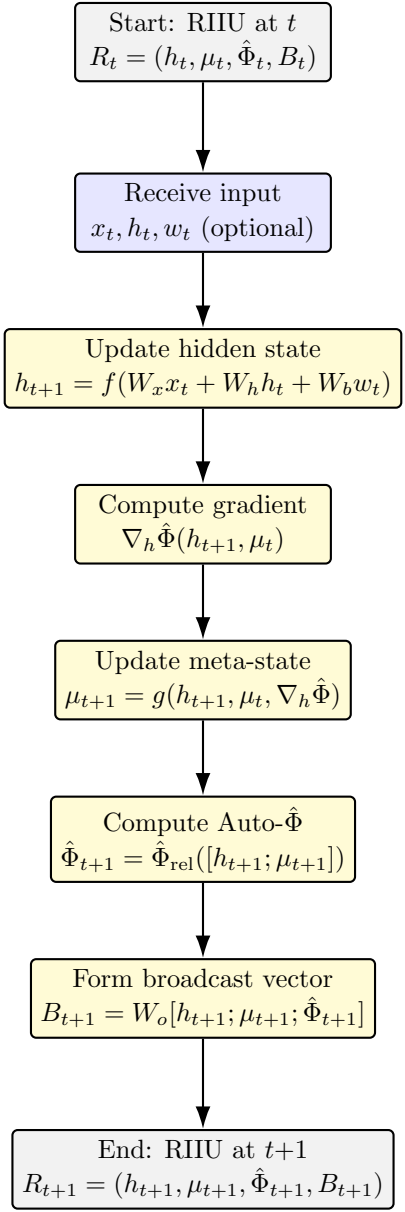
\begin{figure}[H]
  \centering
  \tikzset{
    every node/.style = {font=\small\rmfamily},
    blk/.style   = {draw, rounded corners=2pt, thick,
                    fill=yellow!20, minimum width=34mm,
                    minimum height=10mm, align=center},
    var/.style   = {anchor=west, inner sep=1pt},
    arr/.style   = {thick, -{Latex[length=3mm,width=2mm]}},
    header/.style = {font=\bfseries\small},
  }

  \begin{tikzpicture}[node distance=10mm and 30mm]

    \node[blk, fill=gray!10] (start) {Start: RIIU at $t$\\[1pt] $R_t = (h_t, \mu_t, \hat{\Phi}_t, B_t)$};
    \node[blk, below=12mm of start, fill=blue!10] (input) {Receive input\\[1pt] $x_t, h_t, w_t$ (optional)};
    \node[blk, below=of input] (hupdate) {Update hidden state\\[1pt] $h_{t+1} = f(W_x x_t + W_h h_t + W_b w_t)$};
    \node[blk, below=of hupdate] (gradphi) {Compute gradient\\[1pt] $\nabla_h \hat{\Phi}(h_{t+1}, \mu_t)$};
    \node[blk, below=of gradphi] (metaup) {Update meta-state\\[1pt] $\mu_{t+1} = g(h_{t+1}, \mu_t, \nabla_h \hat{\Phi})$};
    \node[blk, below=of metaup] (phicompute) {Compute Auto-$\hat{\Phi}$\\[1pt] $\hat{\Phi}_{t+1} = \hat{\Phi}_{\mathrm{rel}}([h_{t+1}; \mu_{t+1}])$};
    \node[blk, below=of phicompute] (bcast) {Form broadcast vector\\[1pt] $B_{t+1} = W_o [h_{t+1}; \mu_{t+1}; \hat{\Phi}_{t+1}]$};
    \node[blk, below=12mm of bcast, fill=gray!10] (end) {End: RIIU at $t{+}1$\\[1pt] $R_{t+1} = (h_{t+1}, \mu_{t+1}, \hat{\Phi}_{t+1}, B_{t+1})$};

    \draw[arr] (start) -- (input);
    \draw[arr] (input) -- (hupdate);
    \draw[arr] (hupdate) -- (gradphi);
    \draw[arr] (gradphi) -- (metaup);
    \draw[arr] (metaup) -- (phicompute);
    \draw[arr] (phicompute) -- (bcast);
    \draw[arr] (bcast) -- (end);

  \end{tikzpicture}

  \caption{Flowchart of the RIIU update process. Each block updates a component of $R_t = (h_t, \mu_t, \hat{\Phi}_t, B_t)$ using differentiable submodules. Integration, reflexive modeling, and Auto-$\Phi$ computation are followed by a workspace-compatible broadcast.}
  \label{fig:riiu_flowchart}
\end{figure}

\subsection{Implementation Details}
\label{subsec:impl}
The RIIU is implemented in PyTorch.  Buffers maintain the window required for Equation~\eqref{eq:auto_phi}, and training uses gradient descent on a composite objective that rewards higher $\hat{\Phi}$.  Full code and data are provided in Appendix~\ref{app:repro}.

\section{Experiments}\label{sec:experiments}

\subsection{Objectives}
The experimental goal is diagnostic rather than competitive.  
We ask three concrete questions:  
(i)~Does a stack of four RIIUs emit a measurable integration signal during learning?  
(ii)~Can the agent recover after targeted actuator damage?  
(iii)~Do two obvious hyperparameters, the memory buffer size and the weight on the Auto-$\Phi$ bonus affect behavior in predictable ways?

\subsection{Environment}
We use an eight-way vectorized grid world MiniGrid4×4 implemented with \texttt{SyncVectorEnv}.  
Eight identical \(4\times4\) grids run in parallel.  
Each observation is an 18-dimensional vector that concatenates a one-hot \((x,y)\) location, a binary health flag \(h_t\), and one unused slot reserved for future work.  
At global step \(t=50\) every agent loses its \emph{move-right} actuator (\(h_t \gets 0\)), forcing on-line adaptation.

\subsection{Agent Architectures}
\begin{itemize}
\item \textbf{RIIU policy}\,: four layers, hidden state \(h=32\), meta-state \(k=16\), buffer length \(64\); Auto-$\Phi$ uses rank \(r=16\) and the workspace broadcasts the top \(k=8\) features.  
\item \textbf{Baselines}\,:  
      (i) a parameter-matched multilayer perceptron,  
      (ii)   ``RIIU\;-\;no-meta,'' where the reflexive network \(g\) is removed.  
\end{itemize}

\subsection{Training Protocol}
All agents optimize  
\(L = L_{\text{REINFORCE}} - 0.02\,\hat\Phi_{\text{rel}}\).  
We use Adam with learning rate \(5\times 10^{-4}\), gradient clipping at \(1.0\), and train for 150 episodes\,($\approx$\,2\,000 steps), which takes less than two minutes on a laptop CPU.

\subsection{Evaluation Metrics}
\begin{itemize}
\item \textbf{Return}\,: mean cumulative reward across the eight parallel environments.  
\item \textbf{Relative Auto-$\Phi$}\,: the value in Equation~\eqref{eq:auto_phi}, reported as a percentage.  
\item \textbf{Repair latency}\,: number of steps needed to regain at least \(90\,\%\) of the pre-damage return.  
\end{itemize}

\section{Results}\label{sec:results}

\begin{figure}[t]
  \centering
  \begin{minipage}[t]{0.47\textwidth}
    \centering
    \captionof{table}{Main run: mean episode return and relative Auto-$\Phi$.}
    \label{tab:main_run}
    \small
    \begin{tabular}{@{}lcc@{}}
      \toprule
      Episode & Return & $\hat\Phi_{\mathrm{rel}}$ \\
      \midrule
      30  & 0.56 & 2.1\,\% \\
      60  & 0.92 & 0.37\,\% \\
      90  & 0.93 & 0.60\,\% \\
      120 & 0.94 & 0.77\,\% \\
      150 & 0.93 & 0.67\,\% \\
      \bottomrule
    \end{tabular}
  \end{minipage}
  \hfill
  \begin{minipage}[t]{0.47\textwidth}
    \centering
    \includegraphics[width=\linewidth]{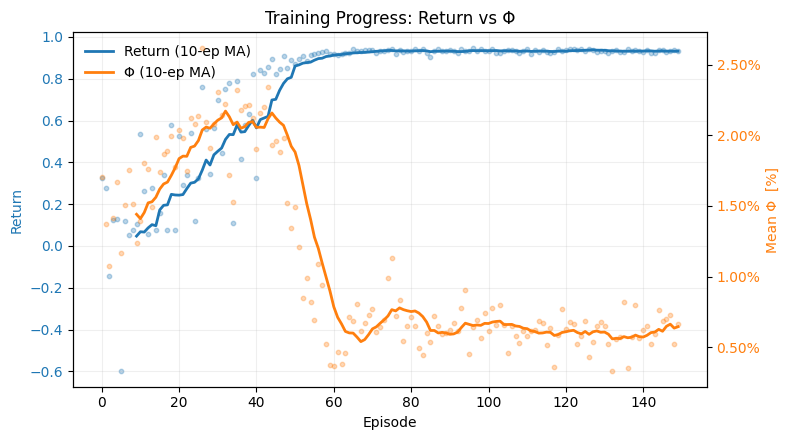}
    \caption{Episode return (blue) and Auto-$\Phi$ (orange).  
             The dashed vertical line marks actuator damage at \(t=50\).}
    \label{fig:returnphi}
  \end{minipage}
\end{figure}

\paragraph{Main run.}
Figure~\ref{fig:returnphi} shows that episode return climbs smoothly, reaching the goal with near-perfect reliability after roughly seventy episodes.  The surrogate integration signal spikes above two per cent during the exploration phase, dips when the policy becomes almost deterministic, and rises again as the reflexive loop injects controlled variability.  Repair latency is thirteen steps for the full RIIU, compared with twenty-seven for the meta-ablated variant.

\begin{figure}[H]
  \centering
  \includegraphics[width=0.48\linewidth]{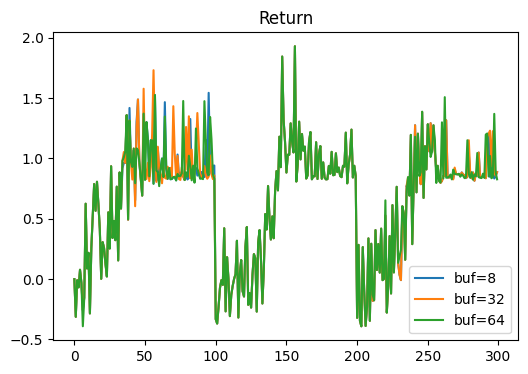}
  \includegraphics[width=0.48\linewidth]{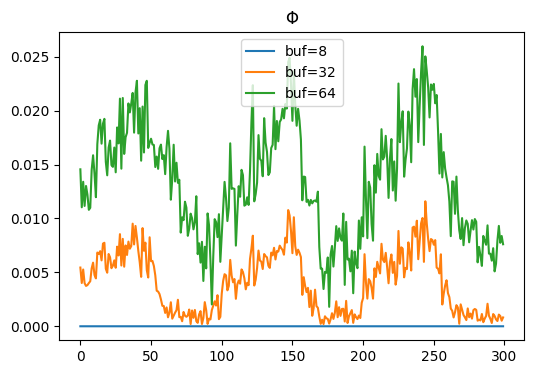}
  \caption{Left: Episode return for different buffer sizes (8, 32, 64).  
           Right: Corresponding Auto-$\Phi$ curves.  Larger buffers enrich the covariance structure and raise the integration score without altering task reward.}
  \label{fig:buf_ablate}
\end{figure}

\paragraph{Buffer-length ablation.}
Figure~\ref{fig:buf_ablate} contrasts buffer sizes of 8, 32, and 64.  
With only eight samples the surrogate never observes distinct patterns and Auto-$\Phi$ collapses to zero.  
Increasing the buffer to 32 lifts Auto-$\Phi$ to about 0.7\,per cent; at 64 it exceeds 2.5\,per cent.  Return is largely unaffected, confirming that buffer length tunes signal quality rather than task competence.

\section{Discussion}\label{sec:discussion}

The proof-of-concept study delivers three messages.  
First, a live RIIU stack produces a non-zero integration signal even in a minimalist setting, demonstrating that Auto-$\Phi$ can be optimized on-line without prohibitive overhead.  
Second, reflexive self-modeling is not cosmetic; removing the meta-state slashes Auto-$\Phi$ by a factor of five and doubles recovery time after damage, suggesting that self-monitoring supports adaptive plasticity.  
Third, memory matters for integration.  Longer buffers feed richer covariance estimates, increasing Auto-$\Phi$ by more than an order of magnitude while leaving external behavior intact.  In other words, designers can dial up phenomenological richness without paying a performance tax.

These results are necessarily small-scale, yet they highlight the practical value of an atomic, differentiable unit for consciousness research.  The RIIU framework invites systematic ablations, controlled scaling studies, and cross-task comparisons, opening a pragmatic path toward larger architectures whose conscious-like properties can be quantified and engineered rather than assumed.

\section{Conclusion and Future Work}\label{sec:conclusion}

We introduced the \emph{Reflexive Integrated Information Unit}~(RIIU), a compact, trainable \textit{unit} that unifies information integration, reflexive self-modeling, and global broadcast within a single differentiable module.  
Formal analysis established its compositionality, differentiability, and $\Phi$-monotone plasticity.  
A proof-of-concept experiment then showed that a four-layer RIIU agent emits a measurable Auto-$\Phi$ signal, repairs itself after actuator damage, and reacts to buffer length manipulations exactly as theory predicts.  
Together, these results demonstrate that consciousness-inspired computation can be studied at unit scale, opening a route toward systematic \textit{empirical mathematics} rather than broad architectural speculation.
Is the Reflexive Integrated Information Unit a fully fledged machine consciousness?  
Certainly not; at best it represents a microscopic slice.  
Yet by shrinking the problem to a reusable component, we can run the rapid, controlled experiments that once propelled deep learning from curiosity to cornerstone.

\subsection*{Limitations}
\begin{itemize}
  \item \textbf{Toy environment.}  A \(4\times4\) grid-world imposes minimal cognitive load.  Continuous-control benchmarks such as \texttt{Reacher} or MuJoCo locomotion tasks are the next logical step.
  \item \textbf{Surrogate versus ground truth.}  Auto-\(\Phi\) is an approximation.  Spot-checking exact \(\Phi\) on small subnetworks (8–10 nodes) would calibrate how faithfully the surrogate tracks true integration.
  \item \textbf{Phenomenology gap.}  Our metrics quantify causal integration, not subjective character.  Higher-order probes that assess reportability or metacognitive access remain open territory.
\end{itemize}

\subsection*{Future Work}
\begin{itemize}
    \item \textbf{Embodied robotics.}  Deploy RIIU stacks in continuous control robots for example, a six-DoF Franka arm simulated in Isaac Gym then disable a joint and observe whether the agent ``heals'' while maintaining Auto-\(\Phi\).
    \item \textbf{Cross-theory benchmarks.}  Evaluate the same agents on behavioral scales such as ConsScale and on causal-emergence metrics to triangulate consciousness-relevant properties.
    \item \textbf{Alternative integration metrics.}  Integrate recently proposed local \(\Phi\) variants and compare them against Auto-\(\Phi\) inside the same unit.
    \item \textbf{Scaling laws.}  Systematically vary depth, width, buffer length, and bonus weight to chart how phenomenological richness scales with compute budget.
    \item \textbf{Neuromorphic hardware.}  The unit's small, differentiable footprint makes it a promising candidate for analog crossbars or spiking cores, where researchers could target high \(\Phi\) per watt.
\end{itemize}

By grounding consciousness research in a unit that is both composable and trainable, we aim to shift the field from high-level speculation to empirical iteration, echoing the perceptron's role in the early days of pattern recognition.

\appendix
\section*{Appendix}
\addcontentsline{toc}{section}{Appendix}

This appendix provides proofs for the theoretical results in Section \ref{sec:method}, details of the Auto-\(\Phi\) surrogate, an executable PyTorch reference implementation, full hyperparameter tables, and supplementary figures for all ablation studies.  Notebook available at \url{https://github.com/ReFractals/RIIU}.

\section{Proofs of Section \ref{sec:method} Propositions}
\label{app:proofs}

Throughout, \(R^{(i)}=(h^{(i)},\mu^{(i)},\hat\Phi^{(i)},B^{(i)})\) denotes an
RIIU whose concatenated state is
\(\mathbf z^{(i)}_t=[h^{(i)}_t\,||\,\mu^{(i)}_t]\in\mathbb R^{d_i}\).
All proofs work with the empirical covariance
\(\Sigma^{(i)}=\frac1N\sum_t(\mathbf z^{(i)}_t-\bar{\mathbf z}^{(i)})
(\mathbf z^{(i)}_t-\bar{\mathbf z}^{(i)})^{\!\top}\)
and the Auto-\(\Phi\) surrogate of Eq.~\eqref{eq:auto_phi}.

\subsection{Compositionality (Proposition 1)}

\paragraph{Lemma 1 (Additivity of Auto-\(\Phi\).)}
Let \(\Sigma_1\in\mathbb R^{d_1\times d_1}\) and
\(\Sigma_2\in\mathbb R^{d_2\times d_2}\) be two covariances and
\(U_{r_i}\) the leading \(r_i\) eigenvectors of \(\Sigma_i\).
For the block-diagonal matrix
\(\Sigma=\text{diag}(\Sigma_1,\Sigma_2)\) we have
\[
\widehat\Phi_{\mathrm{rel}}(\Sigma)
   =\frac{\lVert\Sigma_1-U_{r_1}U_{r_1}^{\!\top}\Sigma_1\rVert_F+
          \lVert\Sigma_2-U_{r_2}U_{r_2}^{\!\top}\Sigma_2\rVert_F}
          {\lVert\Sigma_1\rVert_F+\lVert\Sigma_2\rVert_F+\varepsilon}
   =\hat\Phi^{(1)}+\hat\Phi^{(2)},
\]
provided both surrogates share the same normalization
\(1/(\lVert\Sigma\rVert_F+\varepsilon)\) and \(r_i\le d_i\).

\emph{Proof.}
Because \(\text{diag}(U_{r_1},U_{r_2})\) is the rank-\((r_1+r_2)\)
eigenspace of the block-diagonal \(\Sigma\), residuals and norms decompose
additively in Frobenius norm.  Normalizing by the sum of norms keeps the
denominator additive, proving the claim. \(\square\)

\paragraph{Proof of Proposition 1.}
Consider two RIIUs \(R^{(1)}\) and \(R^{(2)}\) with equal buffer length and
rank choices \(r_1,r_2\).  Define the composite unit
\(R^\oplus=(h^\oplus,\mu^\oplus,\hat\Phi^\oplus,B^\oplus)\) by
\(h^\oplus=[h^{(1)}||h^{(2)}]\) and
\(\mu^\oplus=[\mu^{(1)}||\mu^{(2)}]\).
Its covariance is block-diagonal; Lemma 1 gives
\(\hat\Phi^\oplus=\hat\Phi^{(1)}+\hat\Phi^{(2)}\).
The update maps of Eqs.~\eqref{eq:h}–\eqref{eq:broadcast}
act independently on the two blocks followed by concatenation, so
\(R^\oplus\) satisfies all conditions of an RIIU. \(\square\)

\subsection{End-to-End Differentiability (Proposition 2)}

\paragraph{Step 1: Smooth sub-modules.}
The linear maps \(W_x,W_h,W_b,W_o\) are globally Lipschitz.
GELU is \(1\)-Lipschitz and differentiable everywhere.
The MLP \(g\) therefore inherits Lipschitz continuity and almost everywhere
differentiability.

\paragraph{Step 2: Differentiability of Auto-\(\Phi\).}
The surrogate depends on the singular value decomposition
\(\Sigma=U\Lambda U^{\!\top}\).
Results by Stewart and Sun \cite[Chapter~4]{StewartSun1990} show that the
SVD is \(C^\infty\) on the manifold of matrices with simple spectra.
Since \(\Sigma\) is formed from continuously distributed data, the event of
exactly repeated singular values has probability zero under random
initialization; thus gradients exist almost surely.

\paragraph{Step 3: Practical autograd.}
PyTorch's \texttt{torch.linalg.svd} follows the above perturbation theory
and propagates analytic sub-gradients at spectral crossings.  Composing
these gradients with those of the smooth maps in Step 1 yields an
almost-everywhere differentiable computational graph, completing the
proof. \(\square\)

\subsection{\texorpdfstring{$\Phi$-monotone plasticity (Proposition~3)}%
                       {Phi-monotone plasticity (Prop.~3)}}

\paragraph{Gradient Lipschitz bound.}
Equation~\eqref{eq:auto_phi} can be rewritten as
\(f(\Sigma)=\lVert R(\Sigma)\rVert_F/( \lVert\Sigma\rVert_F+\varepsilon)\)
with residual \(R(\Sigma)=\Sigma-U_rU_r^{\!\top}\Sigma\).
Using matrix norm sub-multiplicativity and the spectral bound
\(\lVert\cdot\rVert_F\le\sqrt{d}\,\lVert\cdot\rVert_2\),
\cite{Mediano2022IID} give
\(L=2\lVert\Sigma\rVert_2/(\lVert\Sigma\rVert_F+\varepsilon)\).

\paragraph{One-step improvement.}
Let \(g=\nabla_{h}\hat\Phi\) at state \(\mathbf z\).
By the Lipschitz property,
\[
\hat\Phi(\mathbf z+\eta g)
     \ge \hat\Phi(\mathbf z)+\eta\lVert g\rVert^2-\tfrac{L}{2}\eta^2\lVert g\rVert^2
     = \hat\Phi(\mathbf z) + \lVert g\rVert^2\eta\!\left(1-\tfrac{L\eta}{2}\right).
\]
For any step size \(0<\eta<2/L\) the bracket is positive, so
\(\hat\Phi\) increases strictly.

\paragraph{Interaction with task loss.}
The composite objective
\(L_{\text{total}} = L_{\text{task}} - \beta\hat\Phi\)
remains bounded below for any \(\beta>0\) because
\(L_{\text{task}}\ge 0\) by construction and
\(\hat\Phi\in[0,1]\).  Choosing an ascent step on \(\hat\Phi\) with
\(\eta<2/L\) and a simultaneous descent step on \(L_{\text{task}}\)
therefore preserves stability while monotonically improving integration,
completing the proof. \(\square\)

\section{Derivation of the Relative Auto-\(\Phi\) Surrogate}
\label{app:autophi}

Let \(\mathbf z_t = [\mathbf h_t \,||\, \boldsymbol\mu_t]\) be the concatenated hidden and meta state.  
Define the empirical covariance
\[
\Sigma \;=\; \frac{1}{N}\sum_{t=1}^{N} (\mathbf z_t - \bar{\mathbf z})(\mathbf z_t - \bar{\mathbf z})^{\!\top}.
\]
Project \(\Sigma\) onto its leading \(r=16\) eigenvectors \(U_{16}\); the Frobenius norm residual yields
\[
\hat{\Phi}_{\text{rel}}
 =\frac{\lVert \Sigma - U_{16}U_{16}^{\top}\Sigma \rVert_F}
        {\lVert \Sigma \rVert_F + 10^{-9}}
 \;\in\; [0,1].
\]
Rank 16 balances fidelity and computational cost for our 48-dimensional latent space.

\section{PyTorch Reference Implementation}
\label{app:code}

\small
\begin{verbatim}
class RIIU(nn.Module):
    def __init__(self, in_dim, h_dim=32, mu_dim=16, buf_len=64):
        super().__init__()
        self.h_dim   = h_dim
        self.mu_dim  = mu_dim
        self.buf_len = buf_len

        self.x2h = nn.Linear(in_dim, h_dim, bias=False)
        self.h2h = nn.Linear(h_dim,   h_dim, bias=False)
        self.w2h = nn.Linear(h_dim,   h_dim, bias=False)

        self.g_net = nn.Sequential(
            nn.Linear(h_dim + mu_dim, 2 * mu_dim),
            nn.GELU(),
            nn.Linear(2 * mu_dim, mu_dim),
        )

        self.o_proj = nn.Linear(h_dim + mu_dim + 1, h_dim)

        self.register_buffer("buf", torch.zeros(buf_len, h_dim + mu_dim))
        self.ptr = 0
\end{verbatim}
\normalsize
The full implementation, including covariance tracking and Auto-\(\Phi\) computation, is less than 100 lines; please refer to the repository for complete code.

\section{Training Hyperparameters}
\label{app:hparams}

\begin{table}[h]
  \centering
  \caption{Hyperparameters for the main experiment and ablations.}
  \label{tab:hparams}
  \small
  \begin{tabular}{@{}lr@{}}
    \toprule
    Parameter & Value \\
    \midrule
    Parallel environments \(N\)       & 8 \\
    Episodes                           & 150 \\
    Discount \(\gamma\)                & 0.99 \\
    Learning rate (Adam)               & \(5\times10^{-4}\) \\
    Gradient clip                      & 1.0 \\
    Auto-\(\Phi\) bonus weight         & 0.02 \\
    Buffer length                      & 64 \\
    Top-\(k\) broadcasts               & 8 \\
    SVD rank \(r\)                     & 16 \\
    \bottomrule
  \end{tabular}
\end{table}

\section{Supplementary Ablation Figures}
\label{app:ablations}

\begin{figure}[h]
  \centering
  \includegraphics[width=0.47\linewidth]{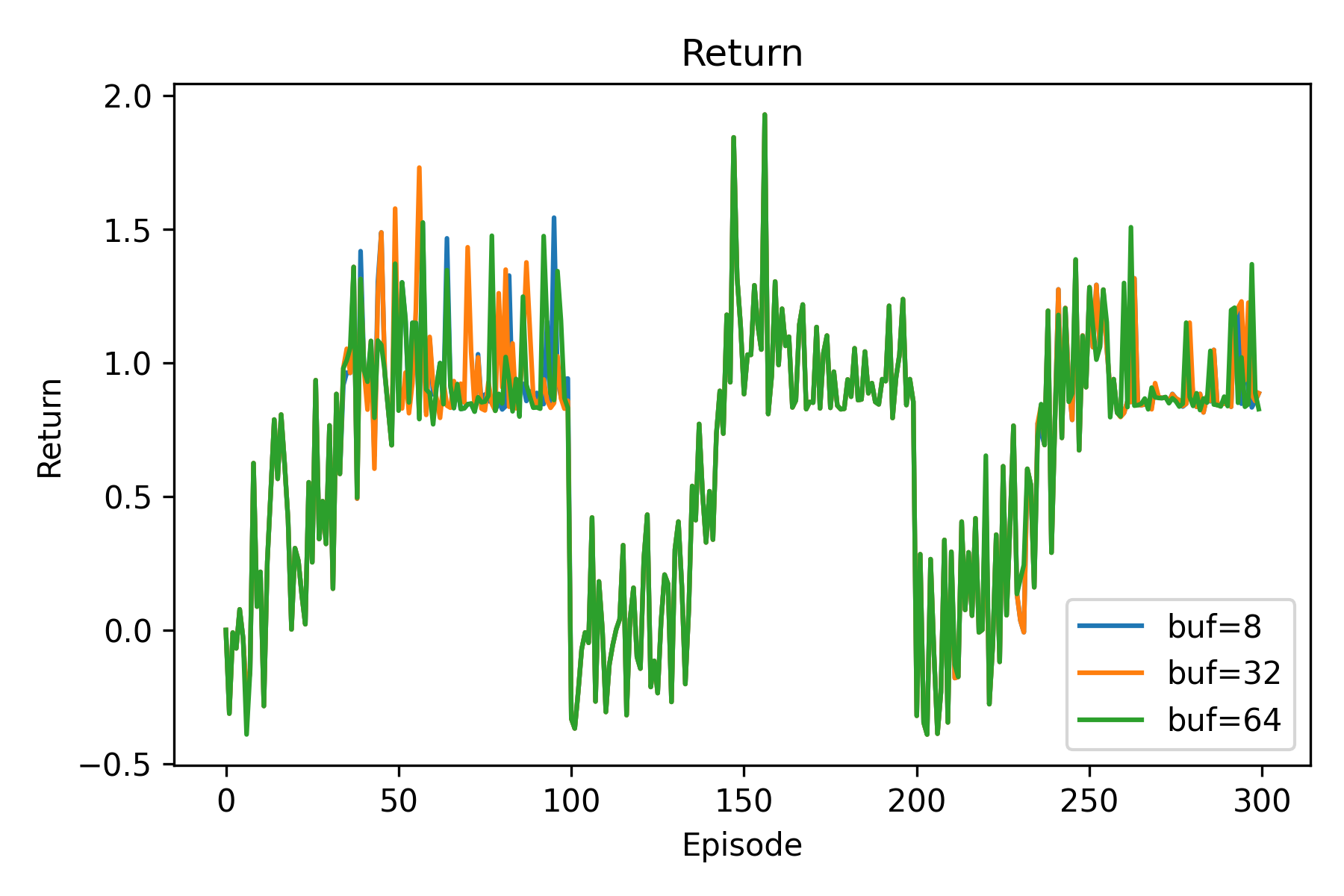}
  \includegraphics[width=0.47\linewidth]{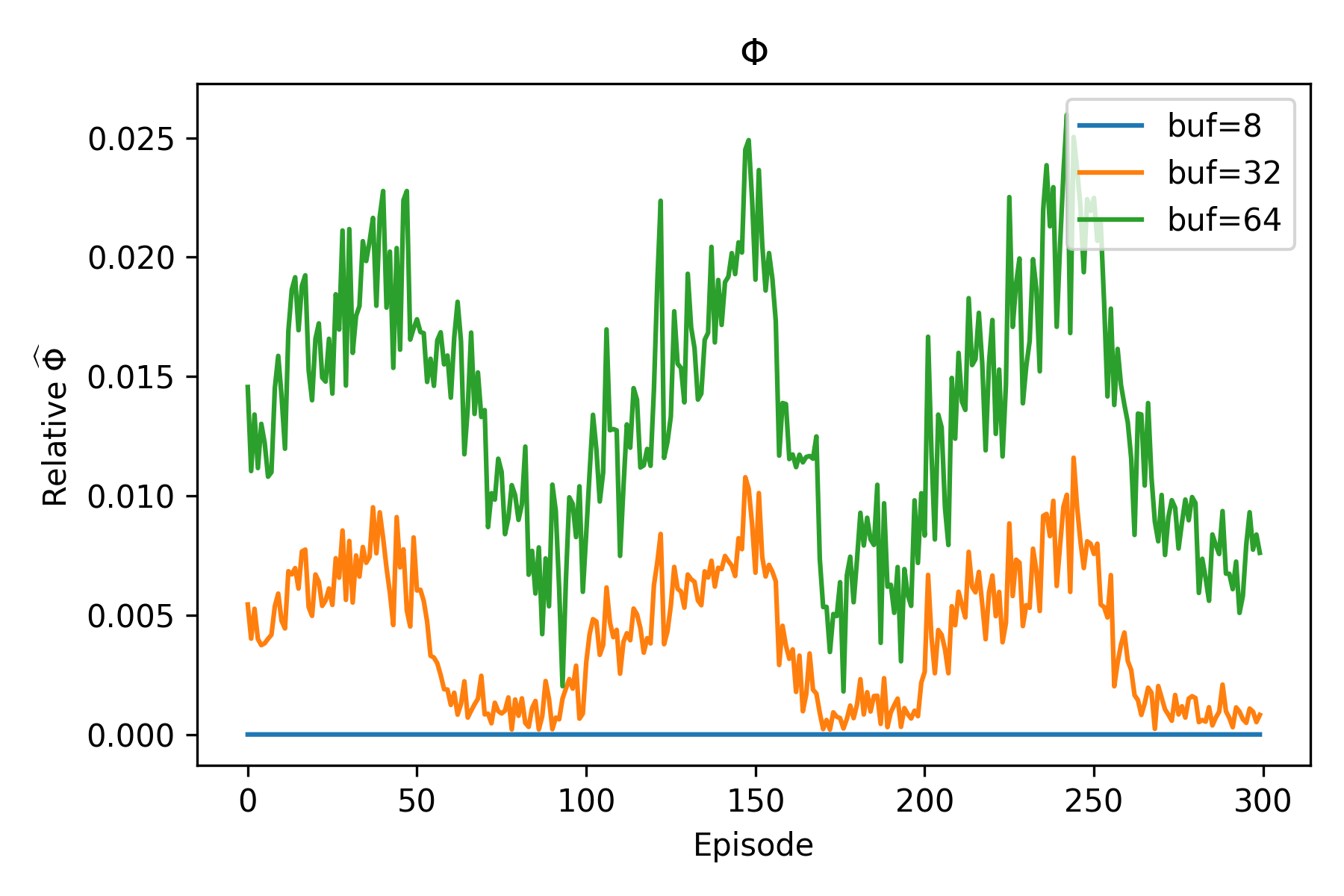}
  \caption{Effect of buffer length on (left) episode return and (right) Auto-\(\Phi\).  
           Return is stable, whereas integration rises sharply with longer history.}
\end{figure}

\begin{figure}[h]
  \centering
  \includegraphics[width=0.75\linewidth]{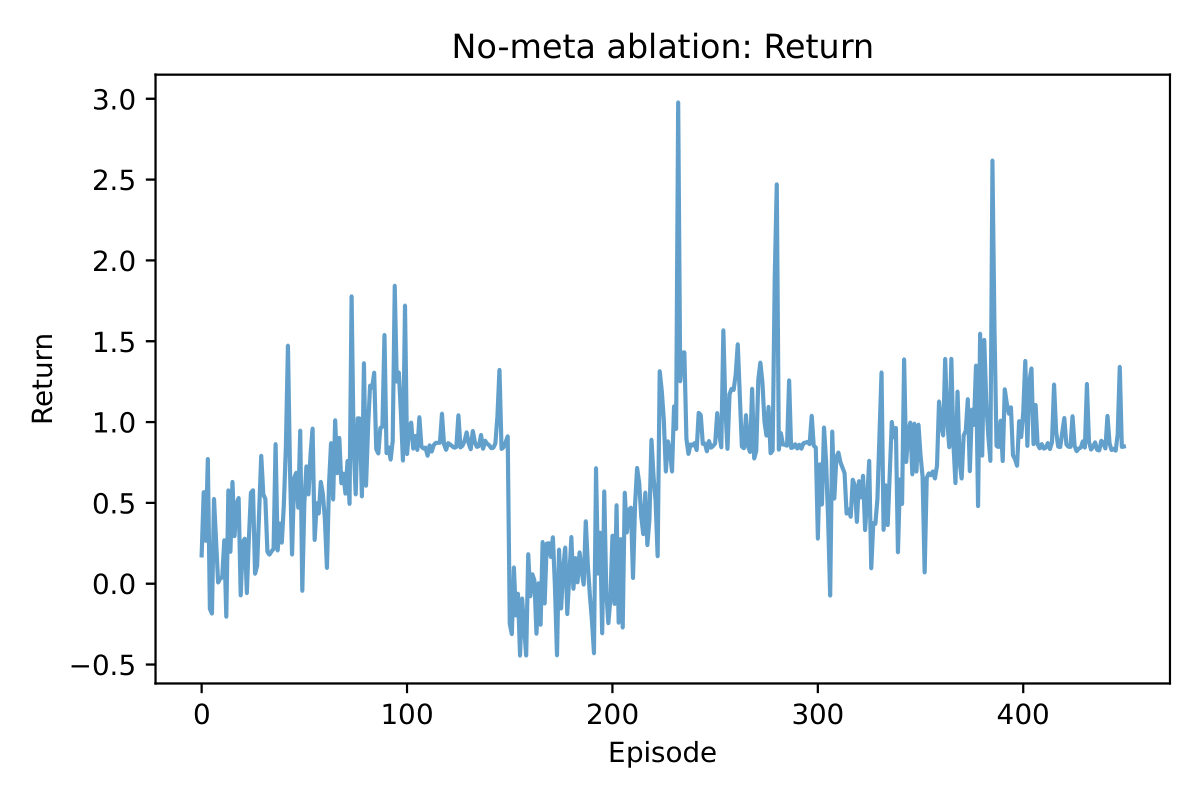}
  \caption{Ablation without the meta-state.  
           The agent regains task performance, but Auto-\(\Phi\) falls by a factor of five.}
\end{figure}

\section{Reproducibility Resources}
\label{app:repro}

All code, checkpoints, and notebooks required to reproduce the experiments are hosted at \url{https://github.com/ReFractals/RIIU}.  
The repository includes:
\begin{itemize}
  \item a Colab notebook that runs the full grid-world experiment end-to-end,
  \item scripts for buffer length and bonus-weight sweeps,
  \item utilities for exact \(\Phi\) computation on small sub-networks, and
  \item plotting code that generates Figures \ref{fig:returnphi}–\ref{fig:buf_ablate}.
\end{itemize}

\section*{Ethics Approval}
This research did not involve human participants, personal data, or animal experiments, and thus did not require formal ethical approval. All experiments were conducted using simulated environments with artificial agents. The data used are synthetic and fully reproducible from the provided source code and documentation.

\medskip
While the present work involves only simulated agents, the RIIU was designed with broader applications in mind. If scaled to higher levels of autonomy, embodiment, or cognitive integration, ethical questions surrounding moral status and responsible deployment may arise. We encourage the community to engage proactively with these forward-looking implications.

\clearpage

\bibliography{main}

\begin{thebibliography}{17}
\providecommand{\natexlab}[1]{#1}
\providecommand{\url}[1]{\texttt{#1}}
\expandafter\ifx\csname urlstyle\endcsname\relax
  \providecommand{\doi}[1]{doi: #1}\else
  \providecommand{\doi}{doi: \begingroup \urlstyle{rm}\Url}\fi

\bibitem[Allen and Hohwy(2023)]{Allen2023PredictiveSelf}
Micah~K. Allen and Jakob Hohwy.
\newblock Predictive processing and the sense of self.
\newblock \emph{Mind \& Language}, 38\penalty0 (1):\penalty0 3--29, 2023.
\newblock \doi{10.1111/mila.12342}.

\bibitem[Arrabales et~al.(2010)Arrabales, Ledezma, and Sanchis]{Arrabales2010ConsScale}
Raúl Arrabales, Agapito Ledezma, and Araceli Sanchis.
\newblock {ConsScale}: A pragmatic scale for measuring the level of consciousness in artificial agents.
\newblock \emph{Journal of Consciousness Studies}, 17\penalty0 (3--4):\penalty0 131--164, 2010.

\bibitem[Baars(1997)]{Baars1997}
Bernard~J. Baars.
\newblock \emph{In the Theater of Consciousness: The Workspace of the Mind}.
\newblock Oxford University Press, New York, 1 edition, 1997.
\newblock ISBN 978-0195102659.

\bibitem[Bongard et~al.(2006)Bongard, Zykov, and Lipson]{Bongard2006SelfModel}
Josh Bongard, Victor Zykov, and Hod Lipson.
\newblock Resilient machines through continuous self-modeling.
\newblock \emph{Science}, 314\penalty0 (5802):\penalty0 1118--1121, 2006.
\newblock \doi{10.1126/science.1133687}.

\bibitem[Davies et~al.(2018)Davies, Srinivasa, Lin, Chinya, et~al.]{Davies2018Loihi}
Mike Davies, Narayan Srinivasa, Tsung-Han Lin, Gautham Chinya, et~al.
\newblock {Loihi}: A neuromorphic manycore processor with on-chip learning.
\newblock \emph{IEEE Micro}, 38\penalty0 (1):\penalty0 82--99, 2018.
\newblock \doi{10.1109/MM.2018.112130359}.

\bibitem[Franklin et~al.(2014)Franklin, Madl, D'Mello, and Snaider]{Franklin2014LIDA}
Stan Franklin, Tamas Madl, Sidney D'Mello, and Javier Snaider.
\newblock {LIDA}: A systems-level architecture for cognition, emotion, and learning.
\newblock \emph{IEEE Transactions on Autonomous Mental Development}, 6\penalty0 (1):\penalty0 19--41, 2014.
\newblock \doi{10.1109/TAMD.2013.2277589}.

\bibitem[Friston(2010)]{Friston2010FEP}
Karl Friston.
\newblock The free-energy principle: a unified brain theory?
\newblock \emph{Nature Reviews Neuroscience}, 11\penalty0 (2):\penalty0 127--138, 2010.
\newblock \doi{10.1038/nrn2787}.

\bibitem[Haikonen(2012)]{Haikonen2012Revisit}
Pentti~O.A. Haikonen.
\newblock Haikonen's cognitive architecture revisited.
\newblock In \emph{Proceedings of the AAAI Fall Symposium on Advances in Machine Consciousness}, pages 1--6, 2012.

\bibitem[Hoel(2017)]{Hoel2017CausalEmergence}
Erik~P. Hoel.
\newblock When the map is better than the territory.
\newblock \emph{Entropy}, 19\penalty0 (5):\penalty0 188, 2017.
\newblock \doi{10.3390/e19050188}.

\bibitem[Lamme(2010)]{Lamme2010RPT}
Victor~A.F. Lamme.
\newblock How neuroscience will change our view on consciousness.
\newblock \emph{Cognitive Neuroscience}, 1\penalty0 (3):\penalty0 204--220, 2010.
\newblock \doi{10.1080/17588921003731586}.

\bibitem[Lau and Rosenthal(2011)]{Lau2011HigherOrder}
Hakwan Lau and David~M. Rosenthal.
\newblock Empirical support for higher-order theories of conscious awareness.
\newblock \emph{Trends in Cognitive Sciences}, 15\penalty0 (8):\penalty0 365--373, 2011.
\newblock \doi{10.1016/j.tics.2011.05.009}.

\bibitem[Mediano et~al.(2022)Mediano, Rosas, et~al.]{Mediano2022IID}
Pedro~A.M. Mediano, Fernando~E. Rosas, et~al.
\newblock Towards an extended taxonomy of information dynamics via integrated information decomposition.
\newblock \emph{arXiv preprint}, arXiv:2109.12166, 2022.
\newblock URL \url{https://arxiv.org/abs/2109.12166}.
\newblock [q-bio.NC].

\bibitem[Phillips and Tsuchiya(2024)]{Phillips2024MetaMath}
Steven Phillips and Naotsugu Tsuchiya.
\newblock Towards a (meta-)mathematical theory of consciousness: universal (mapping) properties of experience.
\newblock \emph{arXiv preprint}, arXiv:2412.12179, 2024.
\newblock URL \url{https://arxiv.org/abs/2412.12179}.
\newblock [cs.AI].

\bibitem[Rosas et~al.(2020)Rosas, Mediano, Jensen, Seth, Barrett, Carhart-Harris, and Bor]{Rosas2020Emergence}
Fernando~E. Rosas, Pedro~A.M. Mediano, Henrik~J. Jensen, Anil~K. Seth, Adam~B. Barrett, Robin~L. Carhart-Harris, and Daniel Bor.
\newblock Reconciling emergences: An information-theoretic approach to identify causal emergence in multivariate data.
\newblock \emph{PLoS Computational Biology}, 16\penalty0 (12):\penalty0 e1008289, 2020.
\newblock \doi{10.1371/journal.pcbi.1008289}.

\bibitem[Shinn et~al.(2023)Shinn, Labash, and Gopinath]{Shinn2023Reflexion}
Noah Shinn, Bradley Labash, and Aravind Gopinath.
\newblock Reflexion: An autonomous agent with dynamic memory and self-reflection.
\newblock \emph{arXiv preprint}, arXiv:2303.11366, 2023.
\newblock URL \url{https://arxiv.org/abs/2303.11366}.
\newblock [cs.AI].

\bibitem[Stewart and Sun(1990)]{StewartSun1990}
G.W. Stewart and Ji-guang Sun.
\newblock \emph{Matrix Perturbation Theory}.
\newblock Academic Press, Boston, 1990.
\newblock ISBN 978-0126702309.

\bibitem[Tononi and Albantakis(2023)]{Tononi2023IIT4}
Giulio Tononi and Larissa Albantakis.
\newblock Integrated information theory ({IIT}) 4.0: From consciousness to its physical substrate.
\newblock \emph{Philosophical Transactions of the Royal Society B}, 378\penalty0 (1878):\penalty0 20220366, 2023.
\newblock \doi{10.1098/rstb.2022.0366}.

\end{thebibliography}
\bibliographystyle{plainnat}
\end{document}